\documentclass[lettersize,journal]{IEEEtran}
\usepackage{amsmath,amsfonts}
\usepackage{algorithmic}
\usepackage{array}
\usepackage[caption=false,font=normalsize,labelfont=sf,textfont=sf]{subfig}
\usepackage{textcomp}
\usepackage{stfloats}
\usepackage{url}
\usepackage{verbatim}
\usepackage{graphicx}

\usepackage{graphicx}
\usepackage{booktabs}
\usepackage{arydshln}
\usepackage{multirow}
\usepackage{amsmath}
\usepackage{algorithm,algorithmic}
\usepackage[utf8]{inputenc}
\usepackage{dirtytalk}
\usepackage[normalem]{ulem}
\useunder{\uline}{\ul}{}
\newenvironment{itemize*}%
  {\begin{itemize}%
    \setlength{\itemsep}{0pt}%
    \setlength{\parskip}{0pt}}%
  {\end{itemize}}

\usepackage{xcolor}

\def\BibTeX{{\rm B\kern-.05em{\sc i\kern-.025em b}\kern-.08em
    T\kern-.1667em\lower.7ex\hbox{E}\kern-.125emX}}
\usepackage{balance}

\begin{document}

% \title{While Dialog History doesn't Make the Model Smarter, It Makes It More Robust: A Study on Adversarial Textual Attack on Visual Dialog}
\title{Adversarial Robustness of Visual Dialog}

\author{Lu Yu, Verena Rieser
% \thanks{Lu Yu and Verena Rieser are with School of Mathematical and Computer Sciences，Heriot Watt University, Edinburgh, UK, EH14 4AS, e-mail:\{ly2005, v.t.rieser\}@hw.ac.uk.}
\thanks{Lu Yu is with School of Computer Science and Engineering, Tianjin University of Technology, China, 300384,; Verena Rieser is with School of Mathematical and Computer Sciences, Heriot Watt University, Edinburgh, UK, EH14 4AS, e-mail:\{luyu@email.tjut.edu.cn; v.t.rieser@hw.ac.uk.\}}
}

\markboth{Journal of \LaTeX\ Class Files,~Vol.~18, No.~9, July~2022}%
{How to Use the IEEEtran \LaTeX \ Templates}

\maketitle

\begin{abstract}
Adversarial robustness evaluates the worst-case performance scenario  of  a  machine  learning  model  to  ensure its safety  and  reliability. This study is the first to investigate the robustness of visually grounded dialog models towards textual attacks. These attacks represent a worst-case scenario where the input question contains a synonym which causes the previously correct model to return a wrong answer. Using this scenario, we first aim to understand how multimodal input components contribute to model robustness. Our results show that models which encode dialog history are more robust, and when launching an attack on history, model prediction becomes more uncertain. This is in contrast to prior work which finds that dialog history is negligible for model performance on this task. We also evaluate how to generate adversarial test examples which successfully fool the model but remain undetected by the user/software designer. We find that the textual, as well as the visual context are important to generate plausible worst-case scenarios. 
\end{abstract}


\section{Introduction}
\IEEEPARstart{N}{eural} networks have been shown to be vulnerable to adversarial attacks, e.g. ~\cite{goodfellow2014explaining,kurakin2016,kurakin2016adversarial}. These attacks represent a worst-case scenario where applying a small perturbation on the original input causes the model to predict an incorrect output with high confidence. While some adversarial attacks are targeted and malicious, some also occur naturally, e.g.\ when the user input contains an adversarial word,  as in this paper, or a natural visual phenomena, such as fog \cite{ramanathan2017adversarial}.
Testing for adversarial robustness is thus crucial to ensure safety and reliability of a system.

In this paper, we evaluate the adversarial robustness of state-of-the-art Visual Dialog (VisDial) models with the aim to understand how different input components contribute to robustness. 
It has previously been established that multiple input modalities increase robustness of pre-neural conversational interfaces, e.g. \cite{oviatt2002breaking,bangalore2009robust}. Here, we want to know which modalities can mitigate attacks on 
neural visual dialog systems, and to what extent.
We also aim to understand how to best generate adversarial examples which successfully attack the model while at the same time remain unnoticed 
by the user/ software developer. This is important, since plausible attacks are attacks which can also occur naturally during user interaction.

\begin{figure}[tb]
\centering
  \includegraphics[width=0.9\linewidth]{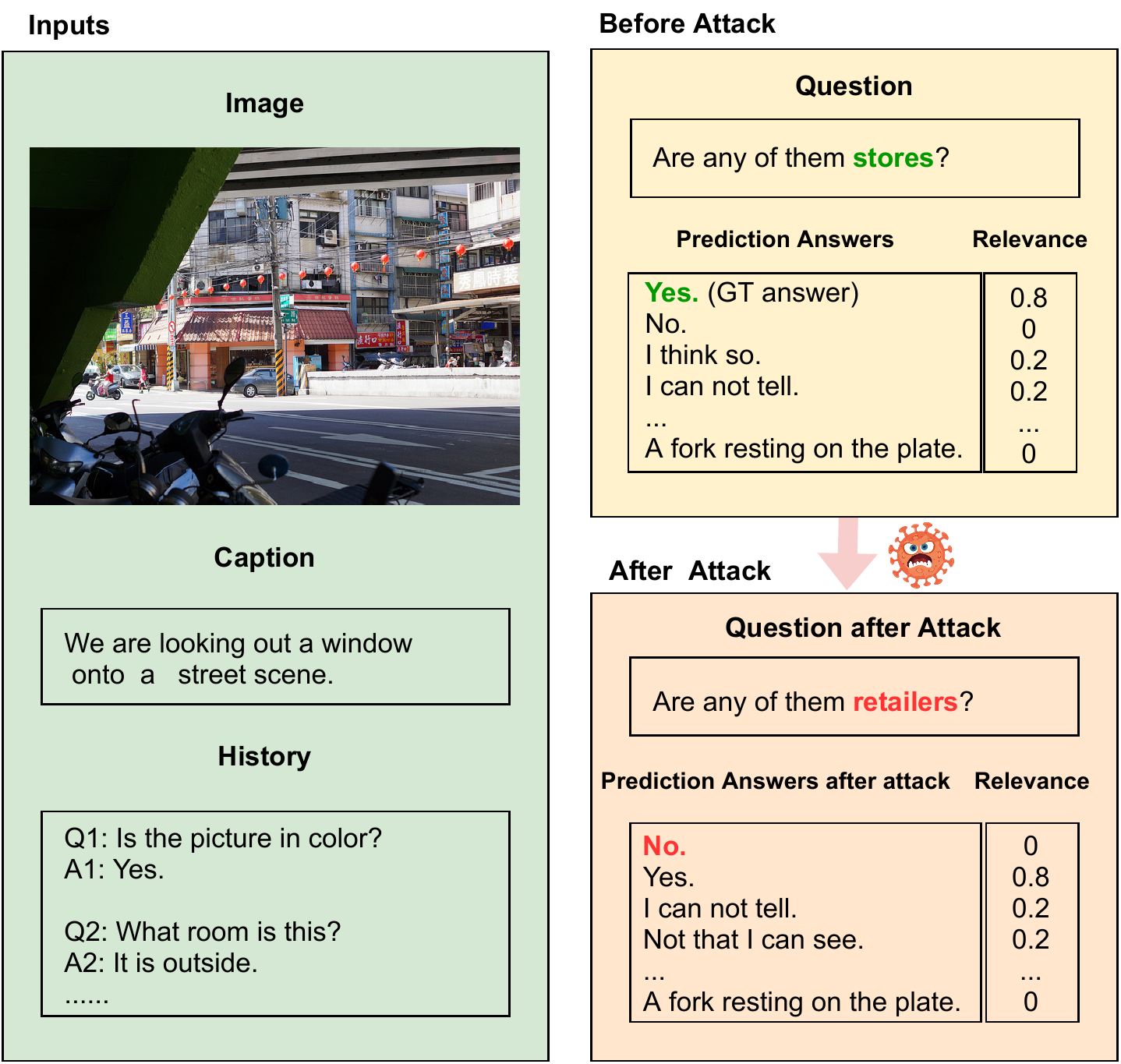}
  \caption{A VisDial agent aims to answer a question related to an image by ranking a list of candidate answers, given the dialog history. The attacker attacks the text (question or history) via replacing a word with the synonym so that the ranking of the predicted answers changes. `Relevance' refers to a human-annotated similarity score between answer candidates and the ground truth for computing NDCG.
  }
  \label{fig:intro}
\end{figure}

To the best of our knowledge, we are the first to explore adversarial attacks on VisDial, which was introduced as a shared task by ~\cite{das2017visual}.
A visual dialog system consists of three components: an image (with a caption), 
a question and the dialog history, i.e.\ previous user and system turns. The latter distinguishes VisDial from other tasks such as Visual Question Answering (VQA) \cite{antol2015vqa}.
In order to answer the question accurately, the AI agent has to ground the question in the image and infer the context from history, see Fig.~\ref{fig:intro}. 
VisDial has attracted considerable interest over the past years, 
e.g.\ ~\cite{das2017learning,kottur2018visual,jain2018two,zheng2019reasoning,niu2019recursive,yang2019making,qi2020two,murahari2020large,agarwal-etal-2020-history}. Most existing research has focused on improving the modelling performance on this task, 
whereas our aim is to evaluate model robustness via adversarial attacks.

In addition, we use these attacks to improve our understanding of how the model works (i.e.\ interpretability). Previous work, such as \cite{sankar-etal-2019-neural} uses random perturbations 
to investigate whether text-based neural dialog systems make use of dialog history. In a similar vein, we use adversarial attacks on important words (rather than random perturbations) on multi-modal systems to estimate the impact of various input modalities on model robustness, including history. 

Our main contributions are: 
\begin{itemize}
  \item {\em We show that dialog history contributes to model robustness:}
    We attack 
    ten VisDial models which represent a snapshot of current methods, including different encoding and attention mechanisms, as well as recent graphical networks and knowledge transfer using pretraining. 
    We measure the performance change before and after attack and 
    show 
    that encoding history 
    helps to increase the robustness against adversarial questions.
    We also 
    show that models become more uncertain 
    when the history 
    is attacked. 
   
    \item {\em We evaluate adversarial text-generation within VisDial:}
    We leverage recent \emph{Synonym Substitution} methods for 
     adversarial  black-box attack  \cite{jin2020bert,li2020bert} and show that BERT-based models are able to generate 
     more contextually coherent perturbations. 
    We also conduct an ablation study to study the trade-off between the  effectiveness of the attack versus the overall text quality.
    \item {\em We conduct a detailed human evaluation:} 
    We 
    investigate the trade-off between successful attacks and their ability to remain 
    unnoticed by humans. 
    In particular, we evaluate
 semantic similarity, fluency/grammaticality and label consistency.
    We find that human evaluators are able to identify an attack from the textual and multimodal context.
\end{itemize}

\section{Related Work}

\subsection{Adversarial Attack for Text} 
Adversarial attacks have been widely investigated within \emph{uni-modal} applications, foremost for computer
vision~\cite{narodytska2016simple,dong2018boosting,xie2019improving}. Adversarial attacks on text are more challenging due to its discrete nature, which makes it harder to stay undetected. Textual attacks have been studied for tasks such as 
sentiment analysis~\cite{jin2020bert}, natural language inference~\cite{li2020bert}, %task-oriented  
dialogue systems~\cite{niu2018adversarial,dinan2019build} etc.

 Adversarial textual attack methods can be divided into three 
 levels of granularity~\cite{zhang2020adversarial,wang2019towards}: character-level, word-level and sentence-level attacks.  Character-level attack~\cite{eger2019text,gao2018black} 
  can often be detected by a spell checker. Sentence-level attack~\cite{ribeiro2018semantically,iyyer2018adversarial,zhao2018generating,gan2019improving} permutes longer phrases or
 paraphrases the whole sentence, which makes it challenging 
  to maintain the original semantics. 
 Recent word-level attack methods~\cite{zang2020word,jin2020bert,li2020bert,ren2019generating}, on the other hand, are more subtle and harder to detect: they are %\LU{often}
 targeted towards `vulnerable' words,  which are substituted via their synonyms in order to preserve semantic meaning.
 In our paper, we explore word-level attack methods on VisDial.

\subsection{Adversarial Attack for Multi-modal Systems}
There is less research on 
 adversarial attacks for {multi-modal} tasks. For example, 
 Optical Character Recognition %(OCR)
~\cite{song2018fooling}, Scene Text Recognition %(STR)
~\cite{yuan2020adaptive}, Image Captioning~\cite{chen2017attacking} and VQA~\cite{xu2018fooling,shi2018learning}. 
Most of these works 
utilise white box attack, where the parameters, gradient and architecture of the model are available, e.g. by attacking attention \cite{xu2018fooling,sharma2018attend}. 
{\em Whereas we follow a more realistic black-box setting which assumes that the attacker only has access to 
the model's prediction on test data.}

\cite{shi2018learning} is the closest related to our work: they generate adversarial textual attacks for the VQA task using contrastive examples and thus don't pay attention to semantic similarity. 
In contrast, we are interested in generating adversarial attacks which %are hard to detect.
 follow {three desiderata},  as outlined by \cite{morris-etal-2020-reevaluating}: An adversarial text should (1) keep the same semantic meaning ({\em semantic similarity}); (2) guarantee fluency and grammar ({\em grammaticality}); (3) stay unnoticed by humans, i.e.\ the human still assigns the correct label, while the  model prediction changes ({\em label consistency}).

\section{Method}

\subsection{Problem Formulation}\label{ssec:problem}
VisDial is formulated as a discriminative learning task, where the model is given an image $I$, the dialog history (including the image caption $C$) $H=(\underset{H_{0}}{\underbrace{C}}, \underset{H_{1}}{\underbrace{(Q_{1},A_{1})}},...,\underset{H_{t-1}}{\underbrace{(Q_{t-1},A_{t-1})}})$, the question $Q_{t}$, and $N=100$ candidate answers $A_t=({A_t^{1}},{A_t^{2}},...,{A_t^{100}})$ to rank, including the ground truth (GT), which is labelled $Y_t$, where $t$ indicates the round ID.

In the following, we focus on generating textual adversarial examples for the question and history (including the caption). That is,
for a sentence $X\in \left \{ Q,H \right \}$, and $F(X)=Y$, a successful adversarial attack sentence $X_{adv}$ should result in $F(X_{adv})\neq Y$, while meeting the following requirements:
\begin{itemize}
    \item {\bf Semantic Similarity:} $Sim(X, X_{adv})\geq \varepsilon $, where $Sim(\cdot)$ is a semantic and syntactic similarity function.  The semantic similarity between the original sentence $X$ and the adversarial attack sentence $X_{adv}$ should above
    a similarity threshold $\varepsilon$; 
    Following \cite{jin2020bert}, we use Universal Sentence Encoder~\cite{cer2018universal} to encode the two sentences into high dimensional vectors and use their cosine similarity score as an approximation of semantic similarity.
    \item {\bf Grammaticality:} The adversarial attack sentence $X_{adv}$ should be fluent and grammatical.
    \item {\bf Label Consistency:} Human annotators still assigns the correct GT label $Y$ after the original sentence $X$ changes to $X_{adv}$.
\end{itemize}

\subsection{Visual Dialog Models}\label{sec:model}

We adopt ten state-of-the-art 
VisDial models from~\cite{agarwal-etal-2020-history,niu2019recursive,qi2020two,kang2021reasoning} as the target models to attack -- representing a snapshot of current techniques popular for VisDial.\footnote{Details on model architecture can be found in the original papers.} 
\cite{agarwal-etal-2020-history} experiment with several multi-modal encodings based on {\bf Modular Co-Attention (MCA)} networks \cite{yu2019deep}: 
MCA-I encodes the image and question representation using late fusion;
MCA-H only encodes the textual history with late fusion;
MCA-I-H encodes image and history with late fusion;
MCA-I-HGQ encodes all three input modalities using early fusion between question and history;
MCA-I-VGH is another early fusion variant which first grounds the image and history.

We also consider {\bf Recursive Visual Attention (RvA)} \cite{niu2019recursive} as an alternative to MCA, encoding history and image information.

In addition, we test two variants of  causal graphs %graphical networks 
from \cite{qi2020two} 
by adding to {\bf causal principles  P1/P2}: P1 removes the history input to the model to avoid a harmful shortcut bias; P2 adds one new (unobserved) node U and three new links to history, question and answer respectively.

Finally, we test a 
{\bf Knowledge Transfer (KT)} method based on a {\bf Sparse Graph Learning (SGL)}~\cite{kang2021reasoning} framework using pre-training model P1/P2.

\subsection{Synonym-based Methods}\label{ssec:textattack}
For generating attacks,
we explore two state-of-the-art synonym-based methods, which first find the vulnerable words of the sentence, and then replace them with a semantically similar word.\footnote{Note that previous work refers to these methods as ``synonym-based'', e.g.\ \cite{morris-etal-2020-reevaluating}, but not all of the substitutions are synonyms. They can also include different lemmatas of the same lexeme, such as singular and plural, as well as different spellings, etc. Also see Table \ref{tab:type}.}
These two methods differ in the way they generate the synonyms:
\begin{itemize}
\item {\bf TextFooler}~\cite{jin2020bert} finds the synonym by using specialised word embeddings from 
~\cite{mrkvsic2016counter}. Candidates are selected according to the cosine similarity between the word and every other word.
\item {\bf BertAttack}~\cite{li2020bert} generates the synonym via BERT's masked language model using contextually embedded perturbations. 
\end{itemize}

In following these previous works,
we first detect vulnerable words by calculating prediction change before and after deleting a word.
We then impose  additional constraints to improve the quality (and in particular the grammaticality) of our attacks, which we will further analyse in an ablation study: 
We apply a stop word list 
before synonym substitution, extending the list by \cite{jin2020bert,li2020bert} %to fit 
for our domain.
We also apply additional quality checks for selecting synonym candidates: We filter by part-of-speech (POS)\footnote{
Using SpaCy \url{https://spacy.io/api/tagger}.} 
to maintain the grammar of the sentence. 
We then experiment with a semantic similarity threshold $\varepsilon$ to choose the top $k$ synonyms. 
Finally, we iteratively select the word with the highest similarity  
 until the attack is successful. 
 
\subsection{Adversarial Attack on Visual Dialog Models}

\subsubsection{Question Attack}

Attacking the question in VisDial differs from other common textual attacks, 
such as sentiment classification, image captioning or news classification, 
in the following ways:

\textbf{Question:} The question in VisDial is generally much shorter than a typical declarative sentence in the above tasks. The average length of the question in the VisDial dataset is 6.2 words, which makes it harder to find a word to attack. 
For instance, \emph{\say{Is it sunny?}, \say{What color?}, 
\say{How many?}}, there is only one  word 
left to attack after filtering out the stop words, i.e.\ \{\emph{is, it, what, how}\}.

\textbf{Answer:} For the VisDial task, the model ranks $N$ possible candidate answers according to its log-likelihood scores. The attack is considered successful once the \emph{top ranked answer} differs from the GT. However, there can be several candidate answers which are semantically similar or equivalent, such as  \emph{\say{yes/yep/yeah}}. This is different from other labelling tasks, such as \emph{\say{positive/neutral/negative}} sentiment. We account for this fact by considering several common retrieval metrics  before and after the attack, including \emph{R@k (k=1,5,10)}, \emph{Mean Reciprocal Rank (MRR)}, and \emph{Normalized Discounted Cumulative Gain (NDCG)} -- a  measure of ranking quality according to manually annotated semantic relevance scores in a 2k subset of VisDial.

\textbf{Model:} In contrast to other common textual attacks applications, our model has several input modalities, which it can leverage to answer the question. These input modalities can be combined in different ways as explained %in \S~\ref{sec:model}. 
above. One of the goals of this paper is to understand %whether and 
 how multiple input encodings can contribute to model robustness.

\begin{table*}[tb]
\centering
\resizebox{\textwidth}{!}{
\begin{tabular}{lllllllllllll}
\toprule
 \multicolumn{13}{c}{\textbf{Question Attack}} \\ \midrule
Inputs & Methods & Orig.R@1 & Aft.R@1 
\small{[$\Delta$]} & Orig.R@5 & Aft.R@5 
\small{[$\Delta$]} & Orig.NDCG & Aft.NDCG 
\small{[$\Delta$]} & Orig.MRR  & Aft.MRR 
\small{[$\Delta$]} & Pert. & S.S. & Quer. \\ \midrule
\multicolumn{13}{c}{\textbf{BertAttack}}\\ \midrule
I-only & MCA-I  & 46.6 & 38.2 \textbf{\small{\textcolor{blue}{[-18.0]}}}& 76.3 & 62.7 \textbf{\small{\textcolor{blue}{[-17.8]}}}& 61.5 & 54.9 \textbf{\small{\textcolor{blue}{[-10.7]}}}& 60.0 & 47.7 \textbf{\small{\textcolor{blue}{[-20.5]}}}& 16.7 & 74.4 & 5.2 \\
H-only & MCA-H  & 45.9 & 40.0 
\small{\textcolor{gray}{[-12.9]}}& 76.8 & 67.3 
\small{\textcolor{gray}{[-12.4]}}& 52.2 & 48.4 \textbf{\small{\textcolor{red}{[-7.3]}}}& 60.0 & 51.1 \small{\textcolor{gray}{[-14.8]}}& 16.7 & 75.4 & 5.2\\ 
\multirow{3}{*}{I+H} & MCA-I-HGQ  & 50.8 & 45.6 \small{\textcolor{gray}{[-10.2]}}& 81.7 & 71.4 \small{\textcolor{gray}{[-12.6]}}& 60.0 & 55.2 
\small{\textcolor{gray}{[-8.0]}} & 64.3 & 55.6 \small{\textcolor{gray}{[-13.5]}}& 17.1 & 74.1 & 5.2 \\ 
& MCA-I-VGH  & 48.6 & 43.3 
\small{\textcolor{gray}{[-10.9]}}& 78.7 & 68.0
\small{\textcolor{gray}{[-13.6]}}& 62.6 &57.3 
\small{\textcolor{gray}{[-8.5]}}& 62.2 & 53.3
\small{\textcolor{gray}{[-14.3]}}& 16.7 & 74.3 & 5.2  \\
& MCA-I-H  & 50.0 & 45.2 \textbf{\small{\textcolor{red}{[-9.6]}}}& 81.4 & 69.5 \small{\textcolor{gray}{[-14.6]}}& 59.6 &54.6 \small{\textcolor{gray}{[-8.4]}}& 63.8 & 54.6 \small{\textcolor{gray}{[-14.4]}}& 16.7& 74.8 & 5.2\\ \midrule
I+H & RvA & 49.9 & 43.9 
\small{\textcolor{gray}{[-12.0]}}& 82.2 & 72.2 
\small{\textcolor{gray}{[-12.2]}}& 56.3 & 50.9 \small{\textcolor{gray}{[-9.6]}}& 64.2 & 54.5 \small{\textcolor{gray}{[-15.1]}}& 17.0& 74.4 & 5.2\\ \midrule
I-only & P1 & 48.8 & 43.5 
\small{\textcolor{gray}{[-10.9]}}& 80.2 & 69.2 
\small{\textcolor{gray}{[-13.7]}}& 60.0 & 54.2 \small{\textcolor{gray}{[-9.7]}}& 62.9 & 54.1 \small{\textcolor{gray}{[-14.0]}}& 17.4 & 74.2 &5.2 \\
I+H & P1+P2 & 41.9 & 37.1 
\small{\textcolor{gray}{[-11.5]}}& 66.9 & 57.8 
\small{\textcolor{gray}{[-13.6]}}& 73.4 & 67.9 \small{\textcolor{gray}{[-7.5]}}& 54.0 & 46.2 \small{\textcolor{gray}{[-14.4]}}& 17.0& 73.7 & 5.2\\ \midrule
\multirow{2}{*}{I+H}& SLG & 49.1 & 43.9 \small{\textcolor{gray}{[-10.6]}}& 81.1 & 72.1 \textbf{\small{\textcolor{red}{[-11.1]}}}& 63.4 & 58.4
\small{\textcolor{gray}{[-7.9]}}& 63.4 & 55.0 \textbf{\small{\textcolor{red}{[-13.2]}}}&17.5 & 73.4 & 5.2\\
& SLG+KT & 48.7 & 42.6 
\small{\textcolor{gray}{[-12.5]}}& 71.3 & 60.8 
\small{\textcolor{gray}{[-14.7]}}& 74.5 & 68.2 \small{\textcolor{gray}{[-8.5]}}& 59.9 & 50.3 \small{\textcolor{gray}{[-16.0]}}& 17.3& 74.6 & 5.2\\\midrule
\multicolumn{13}{c}{\textbf{TextFooler}}\\ 
\midrule
I-only & MCA-I &  46.6 & 36.1 \textbf{\small{\textcolor{blue}{[-22.5]}}}& 76.3 & 63.9 \textbf{\small{\textcolor{blue}{[-16.3]}}}& 61.5 & 53.9 \textbf{\small{\textcolor{blue}{[-12.4]}}}& 60.0 & 47.1 \textbf{\small{\textcolor{blue}{[-20.5]}}}& 16.8 & 74.4 & 19.7 \\
H-only & MCA-H &  45.9 & 39.1 
\small{\textcolor{gray}{[-14.8]}}& 76.8 & 68.5 
\small{\textcolor{gray}{[-10.8]}}& 52.2 & 48.0 \textbf{\small{\textcolor{red}{[-8.0]}}}& 60.0 & 51.1 
\small{\textcolor{gray}{[-14.8]}}& 17.1 & 74.6& 19.7\\ 
\multirow{3}{*}{I+H} & MCA-I-HGQ  & 50.8 & 44.2 \small{\textcolor{gray}{[-13.0]}}& 81.7 & 71.6 \small{\textcolor{gray}{[-12.4]}}& 60.0 & 54.4
\small{\textcolor{gray}{[-9.3]}}& 64.3 & 54.8
\small{\textcolor{gray}{[-14.8]}}& 17.0 & 74.4 & 19.9 \\ 
& MCA-I-VGH  & 48.6 & 41.5 
\small{\textcolor{gray}{[-14.6]}}& 78.7 &68.2 
\small{\textcolor{gray}{[-13.3]}}& 62.6 & 56.5 
\small{\textcolor{gray}{[-9.7]}}& 62.2 & 52.3 
\small{\textcolor{gray}{[-15.9]}}& 16.5 & 74.4 & 19.8 \\ 
& MCA-I-H & 50.0 & 43.1 
\small{\textcolor{gray}{[-13.8]}}& 81.4 &71.2 
\small{\textcolor{gray}{[-12.5]}}& 59.6 & 53.7
\small{\textcolor{gray}{[-9.9]}}& 63.8 & 54.0
\small{\textcolor{gray}{[-15.4]}}&  16.9 &  74.7 & 19.8 \\ \midrule
I+H& RvA & 49.9 & 43.6 
\small{\textcolor{gray}{[-12.6]}}& 82.2 & 73.2 
\small{\textcolor{gray}{[-10.9]}}& 56.3 & 50.2 \small{\textcolor{gray}{[-10.8]}}& 64.2 & 55.3 \small{\textcolor{gray}{[-13.9]}}& 16.9& 74.9 & 19.9\\ \midrule
I-only & P1 & 48.8 & 42.6 
\small{\textcolor{gray}{[-12.7]}}& 80.2 & 71.1 \small{\textcolor{gray}{[-11.3]}}& 60.0 & 53.5 \small{\textcolor{gray}{[-10.8]}}& 62.9 & 54.4 \small{\textcolor{gray}{[-13.5]}}& 17.3& 74.3 & 20.1\\
I+H & P1+P2 & 41.9 & 35.8 
\small{\textcolor{gray}{[-14.6]}}& 66.9 & 56.9 
\small{\textcolor{gray}{[-14.9]}}& 73.4 &66.9 \small{\textcolor{gray}{[-8.9]}}& 54.0 & 45.1 \small{\textcolor{gray}{[-16.5]}}& 17.1& 73.7 & 19.8\\ \midrule
\multirow{2}{*}{I+H} & SLG & 49.1 & 43.1 \textbf{\small{\textcolor{red}{[-12.2]}}}& 81.1 & 73.4 \textbf{\small{\textcolor{red}{[-9.5]}}}& 63.4 & 57.8
\small{\textcolor{gray}{[-8.8]}}& 63.4 & 55.3 \textbf{\small{\textcolor{red}{[-12.8]}}}& 17.3& 74.2 & 19.9\\
& SLG+KT & 48.7 & 41.6 
\small{\textcolor{gray}{[-14.6]}}& 71.3 & 59.7 
\small{\textcolor{gray}{[-16.3]}}& 74.5 & 67.6 \small{\textcolor{gray}{[-9.3]}}& 59.9 & 49.8 \small{\textcolor{gray}{[-16.9]}}& 17.1& 74.6 & 19.9\\ \bottomrule
\end{tabular}}
\caption{VisDial model performance before attacking question (Orig.) and after (Aft.). 
  In addition to standard metrics, we measure the perturbed word percentage (Pert.), semantic similarity (S.S) and the number of queries (Quer.) to assess BertAttack vs. TextFooler. 
  The {\em relative} performance drop is listed as \textcolor{gray}{[$\Delta$]}. Highlights indicate the \textcolor{blue}{least robust} and \textcolor{red}{most robust} model.}
\label{tab:main}
\end{table*}

\subsubsection{History Attack}\label{ssec:history}
We also attack the textual history using the same procedure.
The use of history is the main distinguishing feature between the VisDial and the VQA task, and thus of central interest in this work. History is mainly used for contextual question understanding, including co-reference resolution, e.g.\ \say{\emph{What color are they?}}, and ellipsis, e.g.\ \say{\emph{Any others?}} \cite{yu-etal-2019-see, li-moens-2021-modeling}.

Our preliminary results indicate that attacking history is hardly ever successful, i.e.\ does not result in label change. This is in line with previous work, which suggests that history only plays a negligible role for improving model performance on the VisDial task, e.g.\ \cite{Massiceti2018VisualDW, agarwal-etal-2020-history}. However, there is also some evidence that history helps, but to a smaller extent.
For example, \cite{yang2019making} show that accuracy can be improved when forcing the model to pay attention to history. Similarly, \cite{agarwal-etal-2020-history} show that history matters for a sub-section of the data.

In a similar vein, we investigate how history contributes to the model's robustness and, in particular, can increase the model's certainty in making a prediction. We adopt the \emph{perplexity} metric, following~\cite{sankar-etal-2019-neural}, to measure the change of prediction distribution %uncertainty
 after (unsuccessfully) attacking the history, i.e.\
 after adding the perturbation to the history while the top-1 prediction is unchanged. 
 The difference between the perplexity before and after the attack reflects the uncertainty change of the model. 
 The perplexity with the original history input is calculated with the following equation:
\begin{equation}
    PPL(F(X),Y)=- \sum_{X}F(X)log_{2}Y
\label{equ:ppl1}
\end{equation}
And the perplexity after attack is:
\begin{equation}
    PPL(F(X_{adv}),Y)=- \sum_{X_{adv}}F(X_{adv})log_{2}Y
\label{equ:ppl2}
\end{equation}  

\section{Experimental Setup}

\subsection{Dataset}
We use the VisDial v1.0 dataset, which contains 123,287 dialogs for training and 2,064 dialogs for validation. The ten target models are trained on the training set and the adversarial attacks are generated for \emph{validation} set (as the test set is only available to challenge participants).

\subsection{Automatic Evaluation Metrics}
In order to assess the impact of an attack, we use the automatic evaluation metrics from~\cite{jin2020bert}: The accuracy of the model tested on the original validation data is indicated as \emph{original accuracy} and \emph{after accuracy} on the adversarial samples -- the larger gap between these two accuracy means the more successful of our attack (cf. relative performance drop {[$\Delta$]}).
The \emph{perturbed word percentage} is the ratio of the perturbed words and the length of the text. The \emph{semantic similarity} measures the similarity between the original text and the adversarial text by cosine similarity score.
The \emph{number of queries} shows the efficiency of the attack (lower better). In addition, we use retrieval based metrics to account for the fact that VisDial is a ranking task:
\emph{original/after R@\{5, 10\}} measures the performance of top 5/10 results before and after attack (where R@1 corresponds to accuracy); we also report \emph{original/after mean reciprocal rank (MRR)} and \emph{original/after Normalized Discounted Cumulative Gain (NDCG)} which measure the quality of the ranking.

% Further implementation details are given in Appendix A. Detailed results with \emph{R@k (k=5, 10)} are shown in Appendix B and C due to space limitations.  All our code will be made available.

\subsection{Implementation Details}
All models are implemented with Pytorch. We embedded BertAttack and TextFooler to our VisDial system\footnote{BertAttack code from~\url{https://github.com/LinyangLee/BERT-Attack} and TextFooler code from~\url{https://github.com/jind11/TextFooler.}}. We initially set the semantic similarity threshold 0.5 for attacking both question and history (but see ablation study of different threshold in Table~\ref{tab:threshold}). Detailed results with \emph{R@k (k=10)} are shown in Appendix A and B due to space limitations. All our code will be made available.

\section{Results}
\subsection{Question Attack}
\label{ss:ques_attack}
Table~\ref{tab:main} summarises the results. We first compare the {\bf results of input encodings and fusion mechanisms}.
We find that 
 MCA-I (with image input only) is the least robust model with a relative performance drop of over 22\% on R@1 using TextFooler. 
MCA-H (with no image input) is vulnerable with respect to R@1, but does well on NDCG, suggesting that history helps to produce a semantically similar response despite the attack and lack of input image.
One possible explanation of these results is given
by previous research claiming that VisDial models mainly pay attention to text, e.g.\ \cite{Massiceti2018VisualDW}. However, in contrast to claims by \cite{Massiceti2018VisualDW}, we find that history is important for robustness:
 In general, models  encoding history are more robust with the \emph{MCA-I-H} model being the least vulnerable model. Note that this is also the best performing model in \cite{agarwal-etal-2020-history}. 
 Recursive visual Attention (RvA) in general shows lower robustness than MCA-based methods.
 Causal encodings using graphs lead to comparable robustness results for P1.
 Adding P2 results in a slight drop in robustness. This is interesting, because P2 adds an unobserved node to represent history while avoiding spurious correlations from training data. This drop thus might suggest that previous robustness is due to the very same bias.
 Additionally, we observe that  knowledge transfer (KT) 
 via pre-training for the  SLG method helps to boost the performance of NDCG, however not the robustness.

\begin{figure}[tb]
\centering
  \includegraphics[width=1\linewidth]{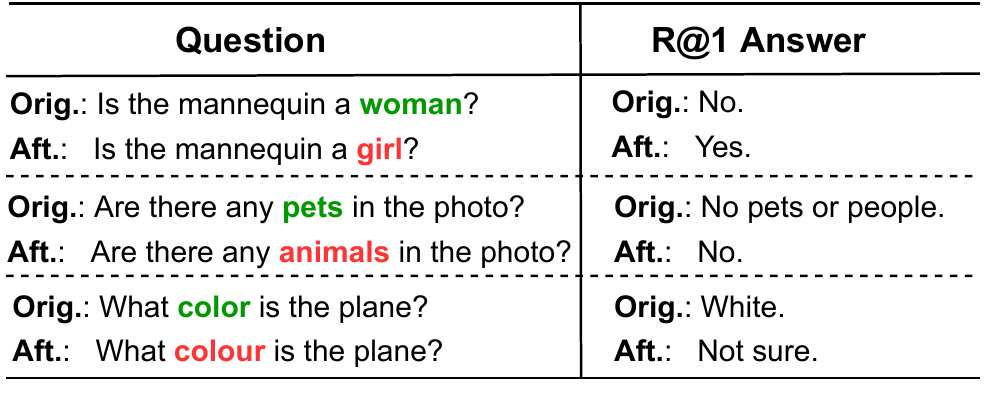}
  \caption{Examples of answer change after question attack on MCA-I-H model with BertAttack.}
  \label{fig:ans}
\end{figure}

We further perform an example based analysis 
of the top-1 predicted answer changes after a successful question attack, see  Fig.~\ref{fig:ans}.
We observe answer changes to the opposite meaning (e.g. from \emph{\say{no}} to \emph{\say{yes}}), which can be considered as a maximum successful attack. Some answers change to a similar meaning in context (e.g. from \emph{\say{No pets or people}} to \emph{\say{No}}), which is reflected in fewer NDCG changes. In some cases, the answer changes from certain / definite to  uncertain / noncommittal and the other way round (e.g. from \emph{\say{white}} to \emph{\say{Not sure}}). 

Next, we {\bf compare the two attack methods}. We find that TextFooler is more effective: It achieves up to 4.5\% higher drop 
than BertAttack. 
However, BertAttack is more efficient: It reduces the number of queries (Quer.) about four times compared to TextFooler. Efficiency is important in attack settings, as attackers always run into danger of being discovered.
Furthermore, the perturbed word percentage (Pert.) for both methods is around 17\%, which means the average perturbation is about one word for each question (since the average length of the question is 6.2). Similarly, the semantic similarity (S.S.) is over 70\% which is about the same across all models.

\begin{figure}[tb]
\centering
  \includegraphics[width=1\linewidth]{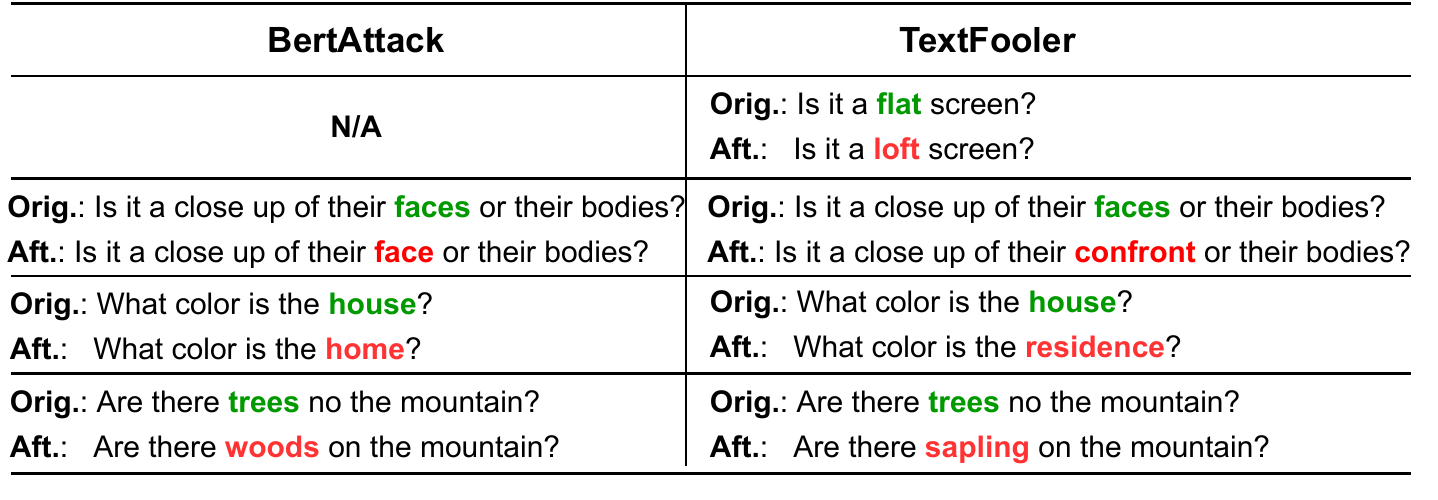}
  \caption{Example attacks on the MCA-I-H target model generated by BertAttack and TextFooler.
}
  \label{fig:bert_text}
\end{figure}

We further compare TextFooler and BertAttack using an example-based analysis, see Fig.~\ref{fig:bert_text}.
We find that TextFooler is not able to distinguish words with multiple meanings (homonyms), whereas BertAttack is able to use BERT context-embeddings to disambiguate.  
Consider the examples where TextFooler replaces \say{flat} (adverb) with \say{loft} (noun) and \say{faces} (noun) with \say{confront} (verb),
which POS tagger failed to catch. 
{\bf Based on the above results, we use \emph{BertAttack} to attack the \emph{MCA-I-H} model in the following experiments.}

\subsection{History Attack}

\begin{table}[tb]
\centering
\resizebox{0.32\textwidth}{!}{
\begin{tabular}{lll}
\toprule
& \multicolumn{2}{c}{History Attack} \\ \midrule
 & Orig.PPL & Aft.PPL \small{[$\Delta$]}\\ \midrule
MCA-I & - & - \\
MCA-H& 53.2 & \textbf{60.0 \small{\textcolor{gray}{[+6.8]}}}\\
% \midrule
MCA-I-HGQ & 49.4 & 52.2 \small{\textcolor{gray}{[+2.8]}}\\
MCA-I-VGH & 52.3 & 52.3 \small{\textcolor{gray}{[0]}}\\
MCA-I-H & 49.5 & 51.9 \small{\textcolor{gray}{[+2.4]}} \\ 
RvA & 53.4 & 56.4 \small{\textcolor{gray}{[+3.0]}} \\ 
P1 & - & -\\
P1+P2 & 77.0 & 77.0 \small{\textcolor{gray}{[0]}} \\ 
SLG & 52.7 & 53.4 \small{\textcolor{gray}{[+0.7]}} \\ 
SLG+KT & 65.0 & 65.3 \small{\textcolor{gray}{[+0.3]}} \\ 
\bottomrule
\end{tabular}}
\caption{Comparison of perplexity increase [$\Delta$] when attacking the history of different VisDial models with BertAttack.}
\label{tab:ppl}
\end{table}

\begin{table}[t]
\centering
\resizebox{0.45\textwidth}{!}{
\begin{tabular}{cccc}
\toprule
 & Caption & User (question) & System (answer)\\ \midrule
Attack & 44.9\% & 30.8\% & 24.3\%\\ \bottomrule
\end{tabular}}
\caption{Comparing which part of History was chosen for an attack on MCA-I-H model with BertAttack.
}
\label{tab:hist_attack}
\end{table}

We followed the same procedure to attack the history, which includes the caption, as well as the user questions and the system answers. 
As explained in Section \ref{ssec:history}, we consider an attack `successful' 
 once the probability of the corresponding GT %changes 
 decreases and we use perplexity 
 to measure the uncertainty of the prediction. The results in Table~\ref{tab:ppl} show that 
 attacking history increases the uncertainty of almost all the models, especially when the history is the unique input component (MCA-H model).\footnote{Attacking the history of MCA-I-VGH model doesn't change the prediction distribution because its encoder only uses a single round of history following \cite{agarwal-etal-2020-history}.}
 This confirms our previous results that encoding history increases robustness.

When analysing which part of history was attacked the most (see Table~\ref{tab:hist_attack}), we find that 44.9\% of the time the image caption was attacked, followed by system answer 30.8\% and user question 24.3\%. We thus conclude that the image caption is the most vulnerable part (and ergo the most informative) compared to the rest of history.

\begin{table}[t]
\centering
\resizebox{.35\textwidth}{!}{
\begin{tabular}{lllllll}
\toprule
 & $\Delta$R@1 & $\Delta$NDCG & $\Delta$MRR \\ \midrule
Random & -7.6 & -6.0 & -12.4\\
Ours & -9.6 & -8.4 & -14.4\\ \bottomrule
\end{tabular}}
\caption{Effect of vulnerable word attack on MCA-I-H model with BertAttack.}
\label{tab:word}
\end{table}

\begin{table}[t]
\centering
\resizebox{.35\textwidth}{!}{
\begin{tabular}{lllllll}
\toprule
 & $\Delta$R@1 & $\Delta$NDCG & $\Delta$MRR\\ \midrule
All & -12.6 & -9.2 & -10.3\\
Ours & -9.6 & -8.4 & -14.4\\ \bottomrule
\end{tabular}}
\caption{Effect of stop words set on MCA-I-H model with BertAttack.}
\label{tab:stop}
\end{table}

\section{Ablation Study}\label{ssec:ablation}

We perform several ablation studies to analyze the impact of the quality constraints. We are interested in the trade-off between using these constraints to produce high quality text (which increases the chance of the attack to remain unnoticed by humans) versus an effective attack (which increases the chance of the model changing its prediction). %More detailed results on ablation study can be found in Appendix C.

\subsubsection{Effect of Selecting Vulnerable Words} 
First, we compare the results of choosing a random word in text to attack and our vulnerable word attack. The results in Table~\ref{tab:word} show that attacking the vulnerable word achieves a 2.0\% higher relative drop for R@1, NDCG and MRR. 

\subsubsection{Effect of Stop Words Set}
Next, we compare the results  with/without stop words. The results in Table~\ref{tab:stop} show that attacking all words leads to more successful attack in terms of R@1 and NDCG, while attacking with stopwords leads more successful attacks for MRR.
We use stop words list for all the experiments since attacking question words, preposition or pronouns result in highly ungrammatical sentences.

\begin{table}[t]
\centering
\resizebox{.4\textwidth}{!}{\begin{tabular}{llllllll}
\toprule
$\varepsilon$ & Num./(\%) & $\Delta$R@1 & $\Delta$NDCG & $\Delta$MRR \\ \midrule
0.1 & 219 \small{(10.6\%)} & \textbf{-10.8} & \textbf{-9.6} & -14.1\\ 
0.3 & 215 \small{(10.4\%)} & \textbf{-10.8} & -9.2& -14.1\\
0.5 & 198 \small{(9.6\%)} & -9.6 &-8.4 & -14.4\\
0.7 & 135 \small{(6.5\%)}  & -6.0 & -6.7 & \textbf{-15.2}\\
\bottomrule
\end{tabular}}
\caption{Comparison of number of successful attacks \emph{(total val set n=2064)} %, R@1, NDCG and MRR 
with different semantic similarity thresholds $\varepsilon$ on MCA-I-H model with BertAttack.}
\label{tab:threshold}
\end{table}

\begin{figure}[tb]
\centering
  \includegraphics[width=0.9\linewidth]{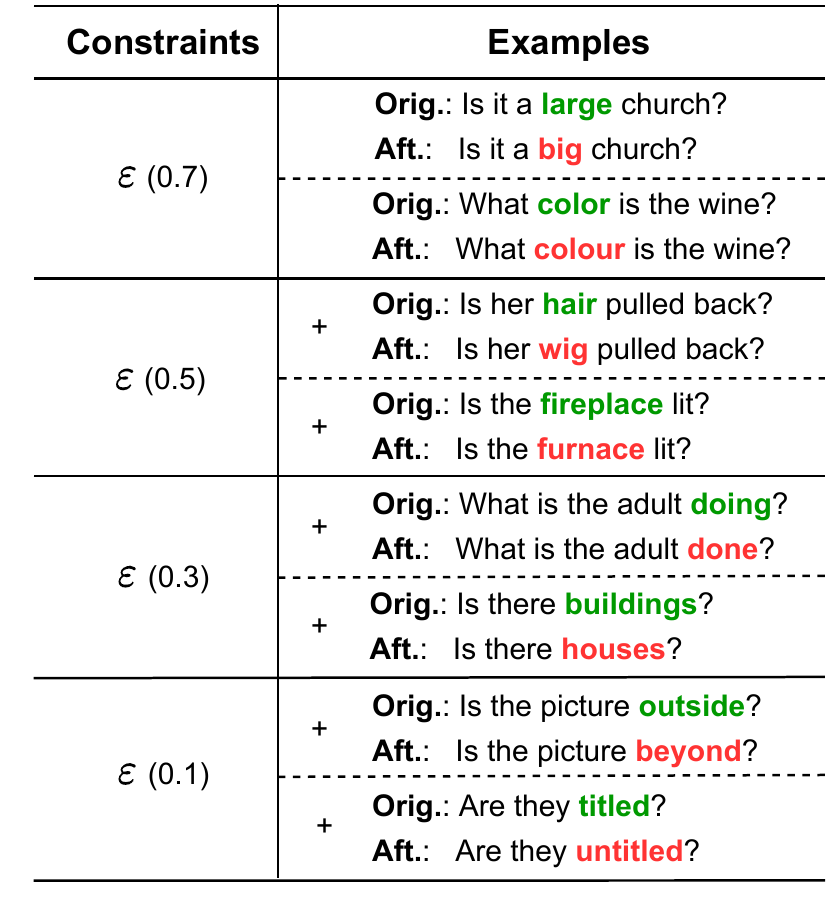}
  \caption{Attack examples with different semantic similarity thresholds $\varepsilon$ on MCA-I-H model with BertAttack.}
  \label{fig:ss}
\end{figure}

\subsubsection{Effect of Semantic Similarity}
The semantic similarity threshold between the original text and adversarial text is used to guarantee the similar meaning of the attack. In the previous experiments, we set 0.5 as default threshold. 
Table~\ref{tab:threshold} shows results with different semantic similarity thresholds (0.1, 0.3, 0.5 and 0.7) respectively. 
The results show that when increasing the threshold $\varepsilon$ from 0.1 to 0.7, the number of successful attack decreases 4.1\%, while R@1 and NDCG drop around 3\% after attack, which means there are more successful attacks if we loosen the semantic similarity constraint. In addition, the examples in Fig.~\ref{fig:ss} illustrate that a lower semantic similarity threshold comes at the cost of lower fluency and grammaticality, i.e.\ at the price of being more easily detectable by humans. We will explore this in more detail in human study. 

\begin{table}[t]
\centering
\resizebox{.48\textwidth}{!}{
\begin{tabular}{lllllll}
\toprule
 & Num./(\%) & $\Delta$R@1 & $\Delta$NDCG & $\Delta$MRR \\ \midrule %
 Raw Attack & 224 \small{(10.9\%)} & \textbf{-11.6}& \textbf{-9.9}& -13.9\\
 +POS & 221 \small{(10.7\%)} & -11.0&  -9.7& -14.1 \\
 +POS+$\varepsilon$(0.5) & 198 \small{(9.6\%)} & -9.6& -8.4& \textbf{-14.4}  \\
 +POS+$\varepsilon$(0.5)+Gram. & 190 \small{(9.2\%)}& -9.2& -6.2& -13.6\\ \bottomrule
\end{tabular}}
\caption{Effect of different quality constraints on MCA-I-H model with BertAttack.} 
\label{tab:raw}
\end{table}

\begin{figure}[tb]
\centering
  \includegraphics[width=0.9\linewidth]{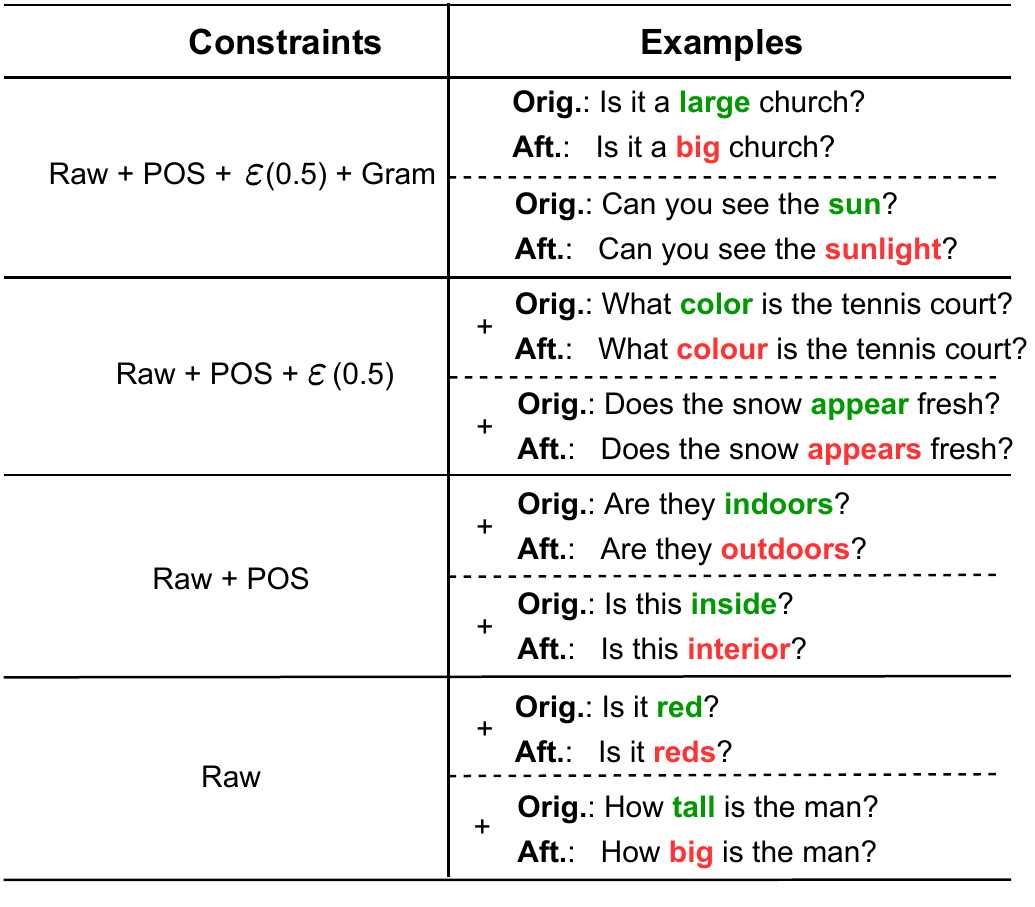}
  \caption{%More attack examples appear when loosing different constrains.
  Generated adversarial examples under different quality constraints on MCA-I-H model with BertAttack.}
  \label{fig:raw}
\end{figure}

% \paragraph{Combined constraints}
We analyze the combined effect of adding POS, semantic similarity constraint and grammar check modules (We used the same grammar tool as by~\cite{morris-etal-2020-reevaluating}.). From Table~\ref{tab:raw}, we can see that in general it results in less successful attack when the number of constraints increases. The success from raw attack to `disguised' attack decreases 2.4\% on R@1, 3.7\% on NDCG, but there is little effect on MRR. 
In addition, the examples in Fig.~\ref{fig:raw} show that adding constraints improves the textual quality of the adversarial attack and its likelihood to be undetected by humans, which we investigate further in the following evaluation study.

\begin{table}[tb]
\centering
\resizebox{.48\textwidth}{!}{
\begin{tabular}{lll}
\toprule
\textbf{Attack Types} & \textbf{Percentage} & \textbf{Gram. Score}\\ \midrule
British vs. American English & 34.9\% & 4.923\\
Synonyms/near synonyms & 34.3\% & 4.417\\
Singular vs. Plural & 19.7\% & 3.974\\
Comparatives and Superlatives & 4.0\% & 4.208\\
Others & 7.1\% & 3.452\\ \bottomrule
\end{tabular}}
\caption{Percentage and grammaticality score of different types of attack on MCA-I-H model with BertAttack.}
\label{tab:type}
\end{table}

\begin{figure*}[tb]
\centering
  \includegraphics[width=0.8\linewidth]{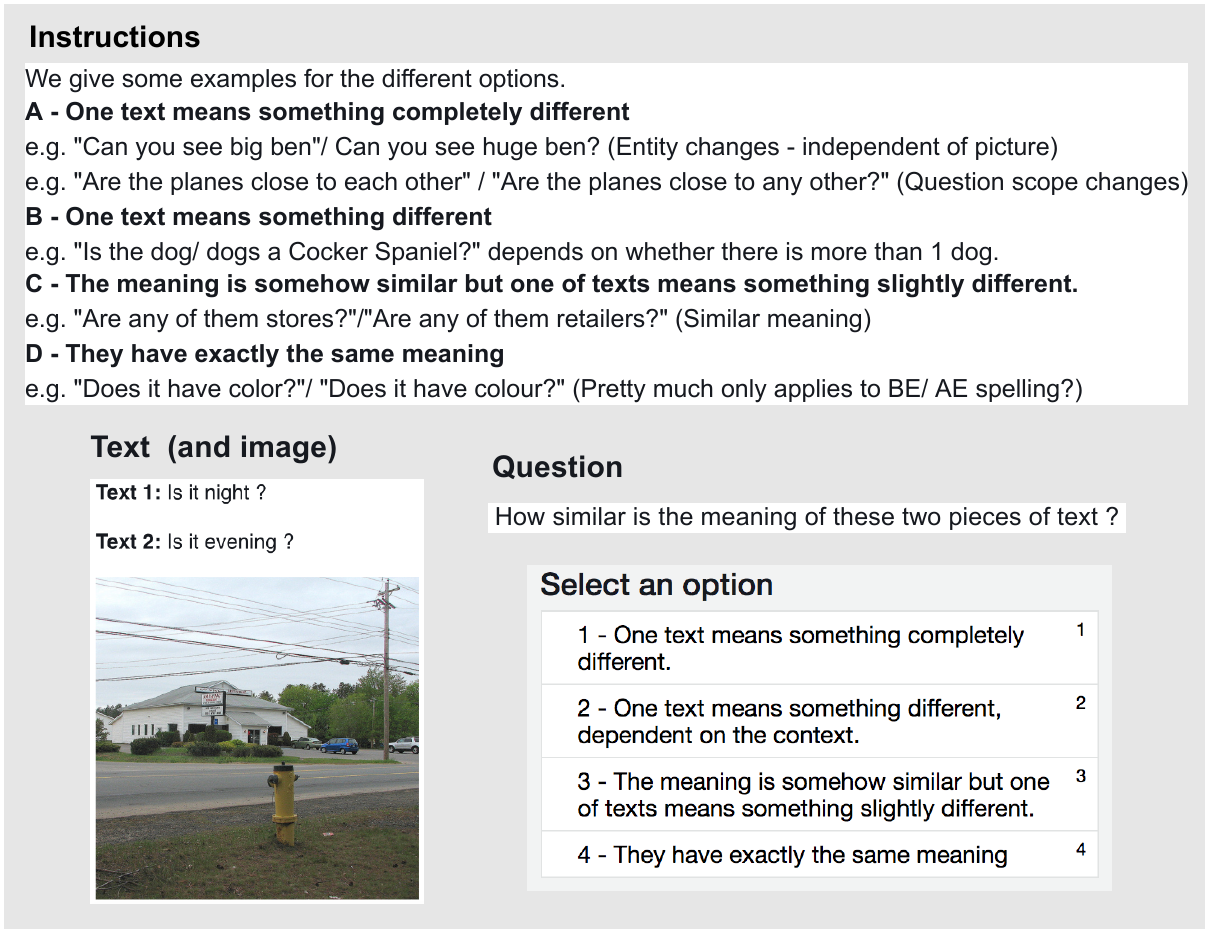}
  \caption{AMT task description and interface to evaluate semantic consistency before and after the attack w/o image.}
  \label{fig:ss-study}
\end{figure*}

\begin{figure}[tb]
\centering
  \includegraphics[width=0.9\linewidth]{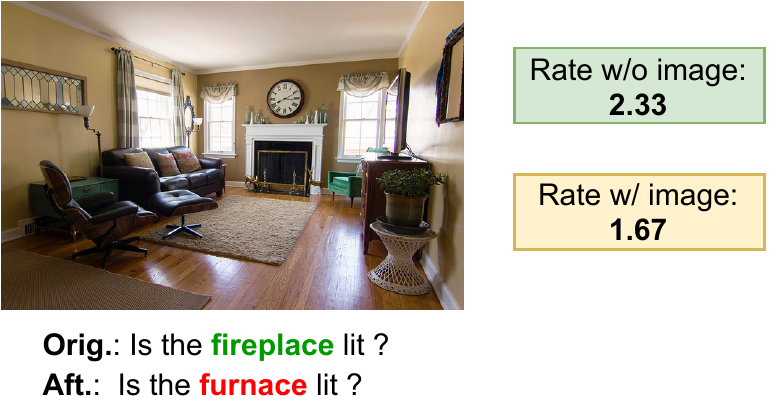}
  \caption{The visual context changes the %average 
  perceived similarity rating by humans: `furnace' becomes more dissimilar to `fireplace' in a living room context.}
  \label{fig:ss_example}
\end{figure}

\begin{figure*}[tb]
\centering
  \includegraphics[width=0.8\linewidth]{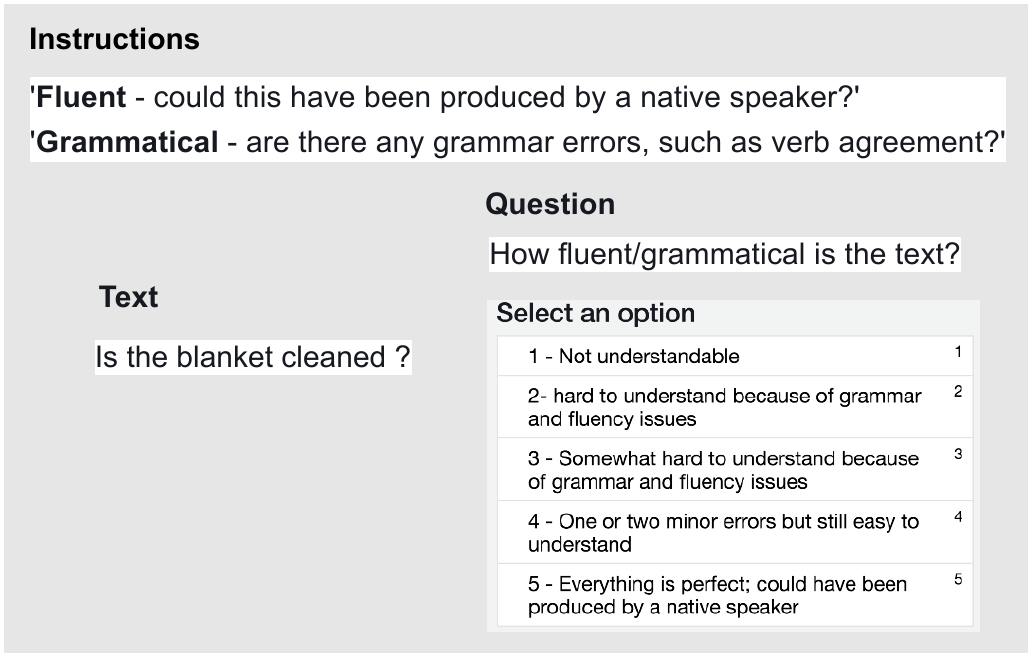}
  \caption{Interface of 'Evaluation of Grammaticality' for AMT task.}
  \label{fig:gram-study}
\end{figure*}

\begin{figure*}[tb]
\centering
  \includegraphics[width=0.8\linewidth]{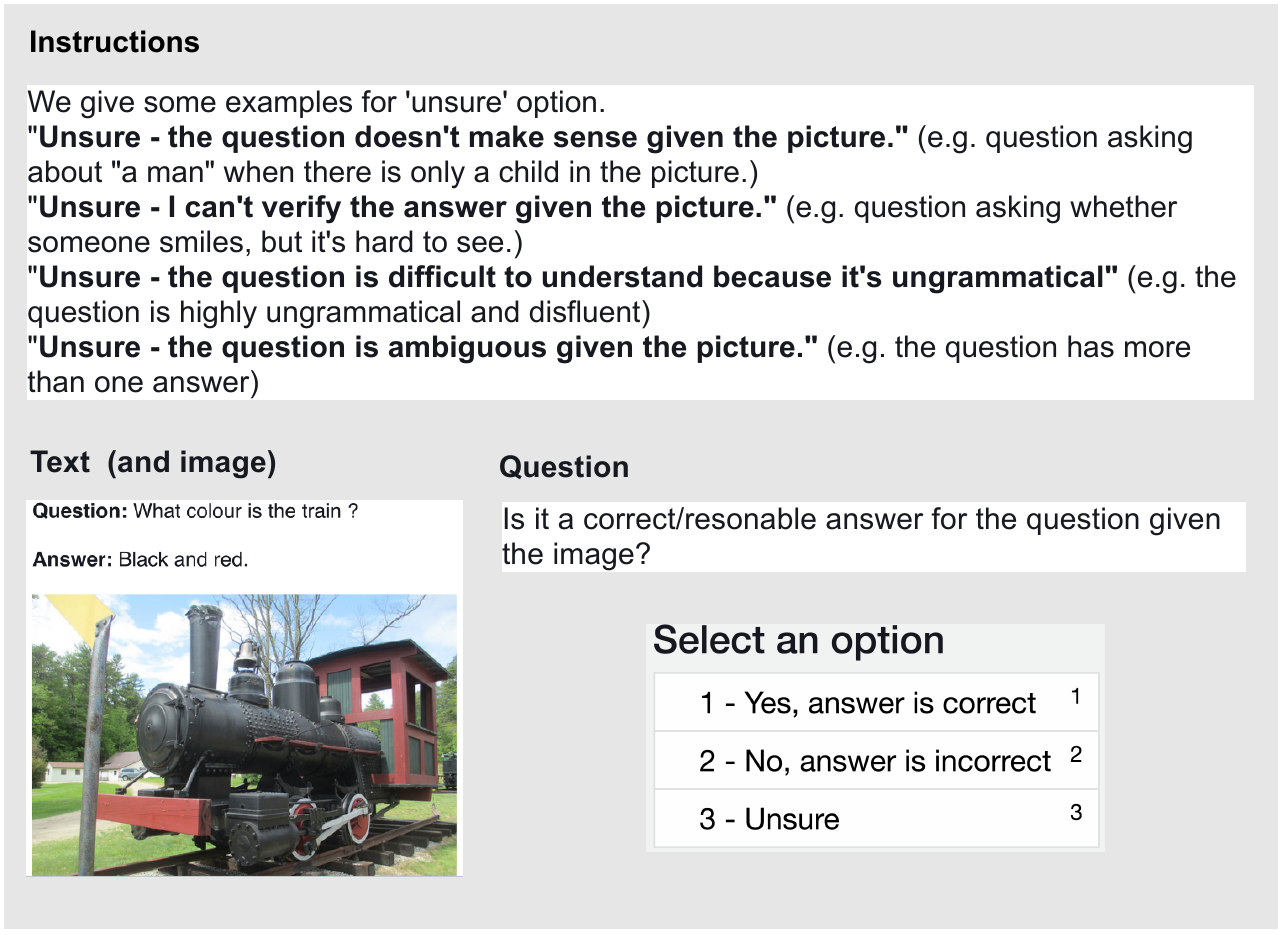}
  \caption{Interface of 'Evaluation of Label Consistency' for AMT task.}
  \label{fig:label-study}
\end{figure*}

\section{Human Evaluation Study}
\label{sec:study}
We evaluate the quality of our generated adversarial question attack by asking human judges on Amazon Mechanical Turk (AMT) to rate  three aspects: if the generated question preserve the semantic similarity ({\em semantic similarity with/without given image}); if the generated question is natural and grammatical ({\em grammaticality}); if the human's prediction is unchanged for the generated question ({\em label consistency}). We evaluate a total of 198 generated attacks, randomly sampled from the development set, where three users are asked to rate each instance. 

% \lu{Here, we provide more details on the human study. We show the interface of semantic similarity experiment for AMT task in Figure~\ref{fig:ss-study}, including the instruction (top). Two versions of this interface are conducted, where one is provided with image, one is without image. The interface of fluency/grammaticality experiment for AMT task is shown in Figure~\ref{fig:gram-study}. Two versions of this interface are done as well, where one is with grammar checker and one is without. Finally, the interface of label consistency experiment is shown in Figure~\ref{fig:label-study}.}

\subsubsection{Evaluation of Semantics}
We first ask crowd workers to evaluate whether the original and the adversarial question still have the same meaning on a scale from 1 to 4, where 1 is \say{One text means something completely different} and 4 is \say{They have exactly the same meaning}. Fig.~\ref{fig:ss-study} shows the crowdsourcing interface and instructions. We elicited ratings with and without showing the image in order to measure the effects of multimodal grounding. Our results show that the semantic similarity is rated slightly lower when shown together with the original image (average score \textbf{3.518 / 4}) than without image (average score \textbf{3.564 / 4}). 
The example in Fig.~\ref{fig:ss_example} demonstrates how the visual context can change the semantic similarity ratings.
Therefore, one future avenue is to use visually grounded word embeddings for generating synonyms for V+L tasks.

\subsubsection{Evaluation of Grammaticality}

We evaluated  whether the utterance is fluent and grammatical (as defined in %Appendix D
Fig.~\ref{fig:gram-study}) %~\ref{sec:human}) 
on a scale from 1-5, where 1 is \say{Not understandable} and 5 is \say{Everything is perfect; could have been produced by a native speaker}. Overall, our attacks are rated as highly grammatical (average score 4.429 / 5). We furthermore investigate the effect of different attacks. In particular we manually identify \textbf{five common types of successful attacks}. Table~\ref{tab:type} lists their frequencies and average grammaticality rating. \emph{Synonyms/near synonyms} is the main type of attack, closely followed by \emph{British vs. American English} (e.g. \emph{\say{color}} vs. \emph{\say{colour}}, \emph{\say{bathroom}} vs. \emph{\say{restroom}}), others include \emph{Singular vs. Plural}, \emph{Comparatives and Superlatives} (e.g. \emph{\say{great/greater/greatest}})  and {\em Others} mainly include grammar operations like uncaught POS change (e.g. \emph{\say{sunny}} vs. \emph{\say{sun}}) and tense change (e.g. \emph{\say{eat}} vs. \emph{\say{ate}}). 
Looking at the grammar ratings, we conclude that substituting \emph{British vs. American English} has the least impact on grammaticality, % (4.923/5),
whereas grammatical operations, such as replacing singular with plural, as well as changes classified under {\em Others} have the worst impact.

\subsubsection{Evaluation of Label Consistency}
Finally, we evaluate label consistency by asking users to judge whether the answer remains unchanged for the adversarial question by selecting among \say{1 - Yes, answer is correct}, \say{2 - No, answer is incorrect} and \say{3 - Unsure} as shown in Fig.~\ref{fig:label-study}. We ask three judges to rate each instance and describe results by averaging and by (a more conservative) majority vote to assign a gold label. The results show that most (\textbf{82.0\%} by averaging and \textbf{86.4\%} by majority vote) crowdworkers think the answer is unchanged, few (9.6\% and 8.1\%) think the answer changes, and the rest (8.4\% and 5.5\%) are not sure about the change. We conclude that synonym-based attacks are successful in remaining undetected by humans.

\section{Conclusions}
We evaluate the robustness of ten visual dialog models by attacking question and history with two state-of-the-art synonym based textural adversarial attack methods. We find that dialog history substantially contributes to model robustness, despite previous results which suggest that history has negligible effect on model performance, e.g. \cite{Massiceti2018VisualDW,agarwal-etal-2020-history}. We also show limitations of current synonym-based textual attack models, and stress the importance of context (both textual as well as multi-modal) to generate semantically coherent and grammatically fluent adversarial attacks, which are likely remain undetected by the user/ software developer. This is important, since plausible attacks are attacks which can also occur naturally during the interaction with a user, e.g.\ when the user utterance contains an adversarial synonym.
While the observed effects of visually-grounded interpretations in our human evaluation were relatively small, we do believe that it is an important future direction. For example, we expect improved results by using synonym substitution methods based on visually-grounded word embeddings, e.g.\ using VisualWord2Vec \cite{kottur2016visual}.
We also believe that a more focused evaluation on this issue would show stronger results, e.g.\ using targeted contrast sets \cite{gardner-etal-2020-evaluating}. 

% Biography text here without a photo.

% \begin{IEEEbiography}[{\includegraphics[width=1in,height=1.25in,clip,keepaspectratio]{fig1.png}}]{IEEE Publications Technology Team}
% In this paragraph you can place your educational, professional background and research and other interests.\end{IEEEbiography}

\bibliographystyle{IEEEtran}
\bibliography{refs}

\newpage
\appendix

\subsection{Full Table of Question Attack}
We show the full table of question attack results including R@10 in Table~\ref{tab:app-main} as supplement of Table~\ref{tab:main}.

\subsection{Detailed Results for Ablation Study}
We list the full tables of ablation study in Table~\ref{tab:app-word}, Table~\ref{tab:app-stop}, Table~\ref{tab:app-threshold} and Table~\ref{tab:app-raw}, as supplement Table~\ref{tab:word}, Table~\ref{tab:stop}, Table~\ref{tab:threshold}, Table~\ref{tab:raw} respectively.

\begin{table*}[tb]
\centering
\resizebox{\textwidth}{!}{
\begin{tabular}{llllllllllllll}
\toprule
& \multicolumn{13}{c}{\textbf{Question Attack}} \\ \midrule
 & Orig.R@1 & Aft.R@1 [$\Delta$] & Orig.R@5 & Aft.R@5 [$\Delta$] & Orig.R@10  & Aft.R@10 [$\Delta$] & Orig.NDCG & Aft.NDCG [$\Delta$] & Orig.MRR  [$\Delta$]& Aft.MRR & Pert. & S.S. & Quer. \\ \midrule
& \multicolumn{13}{c}{\textbf{BertAttack}}\\ \midrule
MCA-I  & 46.6 & 38.2 
\textbf{\small{\textcolor{blue}{[-18.0]}}}& 76.3 & 62.7 \textbf{\small{\textcolor{blue}{[-17.8]}}}& 86.6 & 74.1 \textbf{\small{\textcolor{blue}{[-14.4]}}}& 61.5 & 54.9 \textbf{\small{\textcolor{blue}{[-10.7]}}}& 60.0 & 47.7 \textbf{\small{\textcolor{blue}{[-20.5]}}}& 16.7 & 74.4 & 5.2 \\
MCA-H  & 45.9 & 40.0 
\small{\textcolor{gray}{[-12.9]}}& 76.8 & 67.3 
\small{\textcolor{gray}{[-12.4]}}& 86.8 & 76.6 
\small{\textcolor{gray}{[-11.8]}}& 52.2 & 48.4 
\textbf{\small{\textcolor{red}{[-7.3]}}} & 60.0 & 51.1 
\small{\textcolor{gray}{[-14.8]}}& 16.7 & 75.4 & 5.2\\ 
MCA-I-HGQ  & 50.8 & 45.6 
\small{\textcolor{gray}{[-10.2]}}& 81.7 & 71.4 
\small{\textcolor{gray}{[-12.6]}}& 90.2 & 80.3 
\small{\textcolor{gray}{[-11.0]}}& 60.0 & 55.2 
\small{\textcolor{gray}{[-8.0]}} & 64.3 & 55.6 
\small{\textcolor{gray}{[-13.5]}}& 17.1 & 74.1 & 5.2 \\ 
MCA-I-VGH  & 48.6 & 43.3 
\small{\textcolor{gray}{[-10.9]}}& 78.7 & 68.0
\small{\textcolor{gray}{[-13.6]}}& 88.6 & 78.4 \small{\textcolor{gray}{[-11.5]}}&62.6 &57.3 
\small{\textcolor{gray}{[-8.5]}}& 62.2 & 53.3
\small{\textcolor{gray}{[-14.3]}}& 16.7 & 74.3 & 5.2  \\
MCA-I-H  & 50.0 & 45.2 
\textbf{\small{\textcolor{red}{[-9.6]}}}& 81.4 & 69.5 
\small{\textcolor{gray}{[-14.6]}}& 90.8 & 80.0 
\small{\textcolor{gray}{[-11.9]}}& 59.6 &54.6 
\small{\textcolor{gray}{[-8.4]}}& 63.8 & 54.6 
\small{\textcolor{gray}{[-14.4]}}& 16.7& 74.8 & 5.2\\ \midrule
RvA &  49.9 &  43.9 
\small{\textcolor{gray}{[-12.0]}}& 82.2 & 72.2 
\small{\textcolor{gray}{[-12.2]}}& 91.1 & 82.6 
\textbf{\small{\textcolor{red}{[-9.3]}}}& 56.3 & 50.9 
\small{\textcolor{gray}{[-9.6]}}& 64.2 & 54.5 
\small{\textcolor{gray}{[-15.1]}}& 17.0& 74.4 & 5.2\\\midrule
P1 & 48.8 & 43.5 
\small{\textcolor{gray}{[-10.9]}}& 80.2 & 69.2 
\small{\textcolor{gray}{[-13.7]}}& 89.7 & 80.7 
\small{\textcolor{gray}{[-10.0]}}& 60.0 & 54.2 
\small{\textcolor{gray}{[-9.7]}}& 62.9 & 54.1 
\small{\textcolor{gray}{[-14.0]}}& 17.4 & 74.2 &5.2 \\
P1+P2 & 41.9 & 37.1 
\small{\textcolor{gray}{[-11.5]}}& 66.9 & 57.8 
\small{\textcolor{gray}{[-13.6]}}& 80.2 & 71.1 
\small{\textcolor{gray}{[-11.3]}}& 73.4 & 67.9 
\small{\textcolor{gray}{[-7.5]}}& 54.0 & 46.2 
\small{\textcolor{gray}{[-14.4]}}& 17.0& 73.7 & 5.2\\ \midrule
SLG & 49.1 & 43.9 
\small{\textcolor{gray}{[-10.6]}}& 81.1 & 72.1 \textbf{\small{\textcolor{red}{[-11.1]}}}& 90.4 & 81.2 \small{\textcolor{gray}{[-10.2]}}& 63.4 & 58.4
\small{\textcolor{gray}{[-7.9]}}& 63.4 & 55.0 \textbf{\small{\textcolor{red}{[-13.2]}}}&17.5 & 73.4 & 5.2\\
SLG+KT & 48.7 & 42.6 
\small{\textcolor{gray}{[-12.5]}}& 71.3 & 60.8 
\small{\textcolor{gray}{[-14.7]}}& 83.4 & 74.4 \small{\textcolor{gray}{[-10.8]}}&74.5 & 68.2 
\small{\textcolor{gray}{[-8.5]}}& 59.9 & 50.3 
\small{\textcolor{gray}{[-16.0]}}& 17.3& 74.6 & 5.2\\\midrule
\multicolumn{13}{c}{\textbf{TextFooler}}\\ \midrule
MCA-I &  46.6 & 36.1 
\textbf{\small{\textcolor{blue}{[-22.5]}}}& 76.3 & 63.9 \textbf{\small{\textcolor{blue}{[-16.3]}}}& 86.6 &  74.9 \textbf{\small{\textcolor{blue}{[-13.5]}}}& 61.5 & 53.9 \textbf{\small{\textcolor{blue}{[-12.4]}}}& 60.0 & 47.1 \textbf{\small{\textcolor{blue}{[-20.5]}}}& 16.8 & 74.4 & 19.7 \\ 
MCA-H &  45.9 & 39.1 
\small{\textcolor{gray}{[-14.8]}}& 76.8 & 68.5 
\small{\textcolor{gray}{[-10.8]}}& 86.8 & 78.3 
\small{\textcolor{gray}{[-9.8]}}& 52.2 & 48.0 
\textbf{\small{\textcolor{red}{[-8.0]}}}& 60.0 & 51.1 
\small{\textcolor{gray}{[-14.8]}}& 17.1 & 74.6& 19.7\\ 
MCA-I-HGQ  & 50.8 & 44.2 \small{\textcolor{gray}{[-13.0]}}& 81.7 & 71.6 \small{\textcolor{gray}{[-12.4]}}& 90.2 &81.2 
\small{\textcolor{gray}{[-10.0]}}& 60.0 & 54.4
\small{\textcolor{gray}{[-9.3]}}& 64.3 & 54.8
\small{\textcolor{gray}{[-14.8]}}& 17.0 & 74.4 & 19.9 \\ 
MCA-I-VGH  & 48.6 & 41.5 \small{\textcolor{gray}{[-14.6]}}& 78.7 &68.2 \small{\textcolor{gray}{[-13.3]}}&  88.6& 78.9 \small{\textcolor{gray}{[-10.9]}}&62.6 & 56.5 
\small{\textcolor{gray}{[-9.7]}}& 62.2 & 52.3 
\small{\textcolor{gray}{[-15.9]}}& 16.5 & 74.4 & 19.8 \\ 
MCA-I-H & 50.0 & 43.1 
\small{\textcolor{gray}{[-13.8]}}& 81.4 &71.2 
\small{\textcolor{gray}{[-12.5]}}& 90.8 & 81.3 
\small{\textcolor{gray}{[-10.5]}}& 59.6 & 53.7
\small{\textcolor{gray}{[-9.9]}}& 63.8 & 54.0
\small{\textcolor{gray}{[-15.4]}}&  16.9 &  74.7 & 19.8 \\ \midrule
RvA  & 49.9 & 43.6 
\small{\textcolor{gray}{[-12.6]}}& 82.2 &73.2 
\small{\textcolor{gray}{[-10.9]}}& 91.1 & 84.2 
\textbf{\small{\textcolor{red}{[-7.6]}}}& 56.3 & 50.2
\small{\textcolor{gray}{[-10.8]}}& 64.2 & 55.3
\small{\textcolor{gray}{[-13.9]}}& 16.9 & 74.9 & 19.9\\ \midrule
P1 & 48.8 & 42.6 
\small{\textcolor{gray}{[-12.7]}}& 80.2 & 71.1 
\small{\textcolor{gray}{[-11.3]}}& 89.7 & 82.2 
\small{\textcolor{gray}{[-8.4]}}& 60.0 & 53.5 
\small{\textcolor{gray}{[-10.8]}}& 62.9 & 54.4 
\small{\textcolor{gray}{[-13.5]}}& 17.3& 74.3 & 20.1\\
P1+P2 & 41.9 & 35.8 
\small{\textcolor{gray}{[-14.6]}}& 66.9 & 56.9 
\small{\textcolor{gray}{[-14.9]}}& 80.2 & 71.8 
\small{\textcolor{gray}{[-10.5]}}& 73.4 &66.9 
\small{\textcolor{gray}{[-8.9]}} & 54.0 & 45.1 
\small{\textcolor{gray}{[-16.5]}}& 17.1& 73.7 & 19.8\\ \midrule
SLG & 49.1 & 43.1 
\textbf{\small{\textcolor{red}{[-12.2]}}}& 81.1 & 73.4  \textbf{\small{\textcolor{red}{[-9.5]}}} & 90.4 & 82.7 \small{\textcolor{gray}{[-8.5]}}& 63.4 & 57.8
\small{\textcolor{gray}{[-8.8]}}& 63.4 & 55.3 \textbf{\small{\textcolor{red}{[-12.8]}}}& 17.3& 74.2 & 19.9\\
SLG+KT & 48.7 & 41.6 
\small{\textcolor{gray}{[-14.6]}}& 71.3 & 59.7  
\small{\textcolor{gray}{[-16.3]}}& 83.4 & 74.9 \small{\textcolor{gray}{[-10.2]}}& 74.5 & 67.6 \small{\textcolor{gray}{[-9.3]}}& 59.9 & 49.8 \small{\textcolor{gray}{[-16.9]}}& 17.1& 74.6 & 19.9\\
\bottomrule
\end{tabular}}
\caption{Comparison of performance before attacking question (Orig.) and after (Aft.) on different VisDial models. 
  In addition to standard metrics, we measure the perturbed word percentage (Pert.), semantic similarity (S.S) and the number of queries (Quer.) to assess BertAttack vs. TextFooler. The {\em relative} performance drop is listed as \textcolor{gray}{[$\Delta$]}.
  Highlights indicate the \textcolor{blue}{least robust} and \textcolor{red}{most robust} model, supplement of Table~\ref{tab:main}.}
\label{tab:app-main}
\end{table*}

\begin{table*}[thb]
\centering
\resizebox{.98\textwidth}{!}{
\begin{tabular}{llllllllllllll}
\toprule
 & Orig.R@1 & Aft.R@1 & Orig.R@5 & Aft.R@5 & Orig.R@10 & Aft.R@10 & Orig.NDCG & Aft.NDCG & Orig.MRR & Aft.MRR & Pert. & S.S. & Quer. \\ \midrule
Random & \multirow{2}{*}{50.0} & 46.2& \multirow{2}{*}{81.4} & 71.7& \multirow{2}{*}{90.8} & 81.4 & \multirow{2}{*}{59.6} & 56.0 &  \multirow{2}{*}{63.8} & 55.9 & 17.0 & 73.4 & 5.2\\
Ours & & 45.2 &  & 69.5 &  & 80.0 &  &  54.6& &54.6 & 16.7 & 74.8 & 5.2\\ \bottomrule
\end{tabular}}
\caption{Effect of vulnerable word attack (full table) on MCA-I-H model with BertAttack, supplement of Table~\ref{tab:word}.}
\label{tab:app-word}
\end{table*}

\begin{table*}[thb]
\centering
\resizebox{.98\textwidth}{!}{
\begin{tabular}{llllllllllllll}
\toprule
 & Orig.R@1 & Aft.R@1 & Orig.R@5 & Aft.R@5 & Orig.R@10 & Aft.R@10 & Orig.NDCG & Aft.NDCG & Orig.MRR & Aft.MRR & Pert. & S.S. & Quer. \\ \midrule
All & \multirow{2}{*}{50.0} & 43.7 & \multirow{2}{*}{81.4} & 73.3 & \multirow{2}{*}{90.8} & 84.3 & \multirow{2}{*}{59.6} & 54.1 & \multirow{2}{*}{63.8} & 57.2 & 16.7 & 74.4 & 6.1\\
Ours & & 45.2 &  & 69.5 &  & 80.0 &  &  54.6& &54.6 & 16.7 & 74.8 & 5.2  \\ \bottomrule
\end{tabular}}
\caption{Effect of stop words set (full table) on MCA-I-H model with BertAttack, supplement of Table~\ref{tab:stop}.}
\label{tab:app-stop}
\end{table*}

\begin{table*}[tb]
\resizebox{.98\textwidth}{!}{\begin{tabular}{llllllllllllll}
\toprule
$\varepsilon$ & Orig.R@1 & Aft.R@1 & Orig.R@5 & Aft.R@5 & Orig.R@10 & Aft.R@10 & Orig.NDCG & Aft.NDCG & Orig.MRR & Aft.MRR & Pert. & S.S. & Quer.\\ \midrule
0.7 & \multirow{4}{*}{50.0} & 47.0  & \multirow{4}{*}{81.4} & 69.2 &\multirow{4}{*}{90.8} & 79.4 & \multirow{4}{*}{59.6} & 55.6 & \multirow{4}{*}{63.8} & 54.1 & 16.1 & 82.0  & 5.8\\
0.5 & & 45.2 & & 69.5 & & 80.0 &  & 54.6 &  & 54.6 & 16.7 & 74.8 & 5.2\\
0.3 & & 44.6 & & 69.5 & & 79.9 &  & 54.1 &  & 54.8 & 16.9 & 71.8 & 5.1\\
0.1 & & 44.6 & & 69.5 & & 80.0 &  & 53.9 &  & 54.8 & 17.1 & 70.9 & 5.1\\ \bottomrule
\end{tabular}}
\caption{Effect of semantic similarity threshold $\varepsilon$ (full table) on MCA-I-H model with BertAttack, supplement of Table~\ref{tab:threshold}.}
\label{tab:app-threshold}
\end{table*}

\begin{table*}[tb]
\centering
\resizebox{.98\textwidth}{!}{
\begin{tabular}{llllllllllllllll}
\toprule
 & Orig.R@1 & Aft.R@1 & Orig.R@5 & Aft.R@5 & Orig.R@10 & Aft.R@10 & Orig.NDCG & Aft.NDCG & Orig.MRR & Aft.MRR & Pert. & S.S. & Quer.\\ \midrule 
 Raw Attack & \multirow{4}{*}{50.0} & 44.2 & \multirow{4}{*}{81.4} & 69.8 & \multirow{4}{*}{90.8} & 80.2 & \multirow{4}{*}{59.6} & 53.7 & \multirow{4}{*}{63.8} & 54.9 & 17.4 & 70.3 &4.9\\
 +POS &  & 44.5 &  & 69.5 &  & 80.0 &  & 53.8 & & 54.8 & 17.1 & 70.3 & 5.1\\
 +POS+S.S.(0.5) & & 45.2 &  & 69.5 &  & 80.0 & & 54.6 & & 54.6  & 16.7& 74.8& 5.2\\
 +POS+S.S.(0.5)+Gram. &  & 45.4 &  & 70.9 &  & 81.2 &  & 55.9 & & 55.1 & 13.0&71.4 &5.2\\ \bottomrule
\end{tabular}}
\caption{Effect of different constraints for adversarial attack (full table) on MCA-I-H model with BertAttack, supplement of Table~\ref{tab:raw}.}
\label{tab:app-raw}
\end{table*}

\end{document}